\documentclass[letterpaper, 10 pt, conference]{ieeeconf}  

\IEEEoverridecommandlockouts                              

\usepackage{times}
\usepackage{parskip}
\usepackage[labelfont=bf,textfont=it]{caption}

\usepackage{graphicx}        
\usepackage{url}
\usepackage{hyperref}
\usepackage{amsmath,amssymb}
\usepackage{multirow}
\usepackage[table,xcdraw]{xcolor}
\usepackage{subfigure}
\usepackage{caption}
\usepackage[ruled, vlined, linesnumbered]{algorithm2e}
\usepackage{float}
\usepackage{color}
\usepackage{bm}

\newcommand{\M}[1]{$\mathcal{M}$}

\title{\LARGE 
\textit{The Emotionally Intelligent Robot: }
\\Improving Social Navigation in Crowded Environments}

\author{Aniket Bera$^{1}$, Tanmay Randhavane$^{1}$, Rohan Prinja$^{2}$, \\Kyra Kapsaskis$^{3}$, Austin Wang$^{1}$, Kurt Gray$^{3}$, and Dinesh Manocha$^{4}$
\\
\thanks{$^{1}$Authors from the  Department of Computer Science, University of North Carolina at Chapel Hill, USA}
\thanks{$^{3}$Author from Google, USA}
\thanks{$^{3}$Authors from the Department of Psychology and Neuroscience, University of North Carolina at Chapel Hill, USA}
\thanks{$^{4}$Author from the Department of Computer Science, University of Maryland at College Park, USA}%
}

\definecolor{yellow1}{RGB}{239, 192, 35}
\definecolor{green1}{RGB}{12, 181, 63}
\definecolor{purple1}{RGB}{132, 12, 181}

\begin{document}

\maketitle

\begin{abstract}
We present a real-time algorithm for emotion-aware navigation of a robot among pedestrians. Our approach estimates time-varying emotional behaviors of pedestrians from their faces and trajectories using a combination of Bayesian-inference, CNN-based learning, and the PAD (Pleasure-Arousal-Dominance) model from psychology. These PAD characteristics are used for long-term path prediction and generating proxemic constraints for each pedestrian. We use a multi-channel model to classify pedestrian characteristics into four emotion categories \textit{(happy, sad, angry, neutral)}. In our validation results, we observe an emotion detection accuracy of $85.33\%$. We formulate emotion-based proxemic constraints to perform socially-aware robot navigation in low- to medium-density environments. We demonstrate the benefits of our algorithm in simulated environments with tens of pedestrians as well as in a real-world setting with Pepper, a social humanoid robot.
\end{abstract}

\section{Introduction}
\vspace*{-0.1in}
Recent advances in technology predict that humans will soon be sharing spaces in public places, sidewalks, and buildings with mobile, autonomous robots. Recently, mobile robots are increasingly used for surveillance, delivery and warehousing applications. It is important that such robots navigate in socially acceptable ways, meaning that they seamlessly navigate through pedestrian traffic while responding dynamically -- and appropriately -- to other pedestrians.

\begin{figure}
\centering
\includegraphics[width=0.5\textwidth]{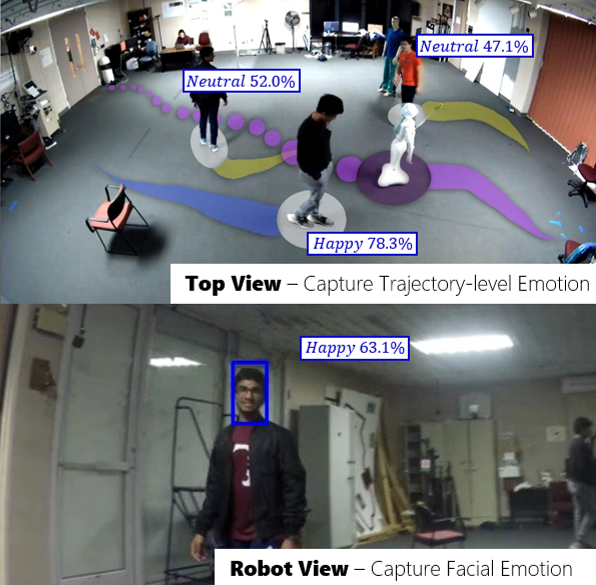}
\vspace*{-0.15in}
\caption{\textbf{The Emotionally Intelligent Robot}: We present a real-time data-driven planning algorithm that learns the emotion state of the pedestrians to perform socially-aware navigation. (Top) The robot learns pedestrians' emotions and their proxemic constraints to improve both social comfort and navigation. (Bottom) The robot extracts facial expressions using an onboard camera and combines it with the trajectory information from the camera in (Top) to efficiently compute the overall emotions of the pedestrians and perform socially-aware navigation.} 
\label{fig:cover}
\vspace{-18pt}
\end{figure}

\begin{figure*}[t]
\centering
\includegraphics[width=1\textwidth]{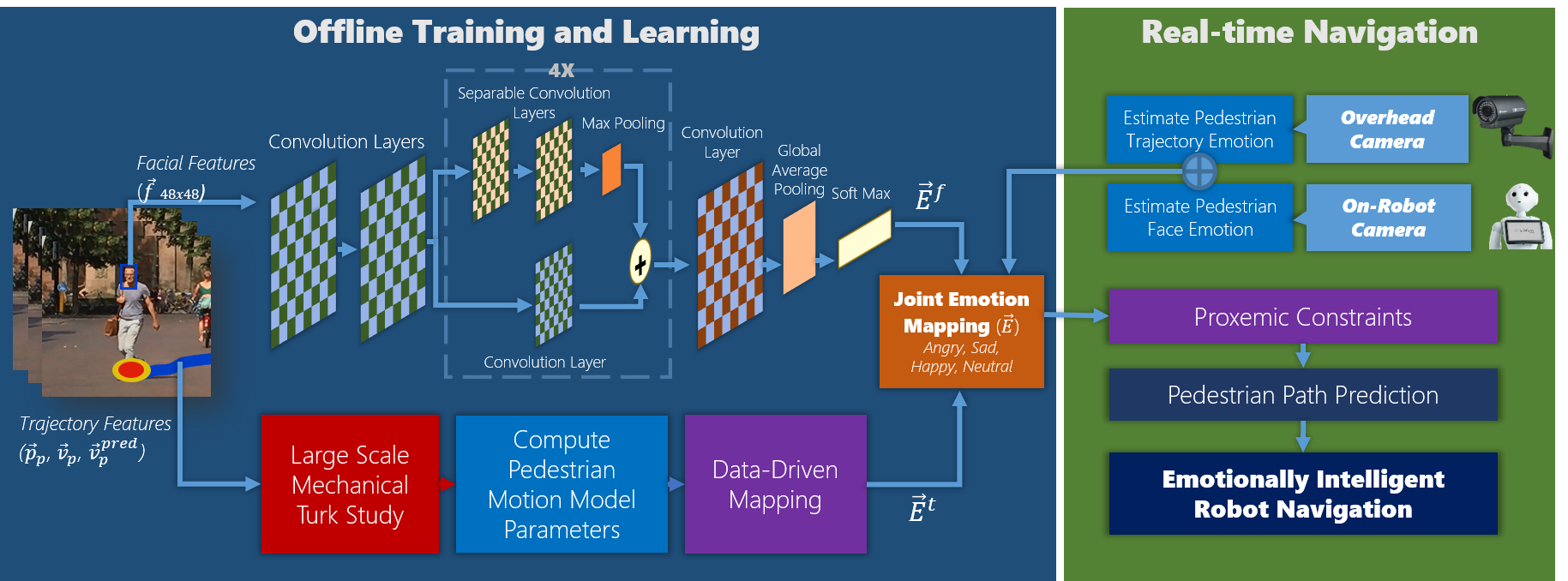}
\caption{\textbf{Overview}: Our method takes a streaming video as input from two channels, 1) fixed, overhead camera and 2) onboard robot camera. We perform a large scale Mechanical Turk study on a crowd dataset to precompute a data-driven mapping between a motion model and their emotions. At runtime, we use this mapping along with the trajectory to compute emotion, $\vec{E}^t$, for the pedestrians. Using 2) we use a fully-convolutional neural network (which has been trained on the FER-2013 emotion dataset~\cite{goodfellow2013challenges}) to compute the emotion based on facial cues, $\vec{E}^f$. We combine these multi-channel emotions with proxemic constraints and a collision-avoidance algorithm to perform socially-aware robot navigation through pedestrians.}
\vspace{-15pt}
\label{fig:overview}
\end{figure*}

A robot navigating through the world alone is primarily a physical problem (compute collision-free and efficient paths that satisfy the kinematics and dynamics constraints of the robot) because it has to make its way around obstacles. However, when there are other pedestrians in this environment, navigation becomes just as much about social navigation as it does about physical navigation. Humans act as both dynamic and social obstacles to a robot and have their own intentions, desires, and goals, which can affect a robot's progress. Additionally, a robot's movement may also affect humans' comfort and/or emotional state.






To predict other people's goals, people use a variety of cues, including past behavior, speech utterances, and facial expressions~\cite{Kilner2007}. One of the most important predictors of people's behavior is their emotions~\cite{Zeelenberg2008}, and therefore understanding people's emotional states is essential for making your way through the social world~\cite{barrett2017}.  The ability to understand people's emotions is called ``emotional intelligence''~\cite{Mayer1990} and is useful in many social situations, including predicting behavior and navigation. As more robots are introduced in social settings, techniques to develop emotional intelligence of robots become increasingly important in addition to merely satisfying the physical constraints.

However, understanding the emotions of pedestrians is a challenging problem for a robot. There has been considerable research on using non-verbal cues such as facial expressions to perceive and model emotions~\cite{fabian2016emotionet}. However, recent studies in the psychology literature question the communicative purpose of facial expressions and doubt the reliability of emotions perceived only from these expressions~\cite{russell2003facialandvocal}. 
There are many situations where facial data is only partially available or where facial cues are difficult to obtain. For example, a pedestrian may not be directly facing the robot or may be far away from the robot. Therefore, combining facial expressions with a more implicit channel of expression such as trajectories is vital for more accurate prediction of humans' emotional states. 

\noindent{\bf \large Main Results:} We present a real-time data-driven planning algorithm that takes the emotional state of the pedestrians into account to perform socially-aware navigation (Figure~\ref{fig:cover}). We predict pedestrians' emotions based on the Pleasure-Arousal-Dominance (PAD) model, a 3-dimensional measure of emotional state used as a framework for describing individual differences in emotional traits/temperament~\cite{Mehrabian1996}, using information from two different channels of expression: faces and trajectories. We extract the trajectory of each pedestrian from a video stream and use Bayesian learning algorithms to compute their motion model and emotional characteristics. This trajectory-based computation is based on the results of a perception user study that provides emotion labels for a dataset of walking videos.  In our validation studies, we observe an accuracy of $85.33\%$ using 10-fold cross-validation.  We also compute the facial expression-based emotion using a convolution-neural network (CNN) classifier trained on the FER-2013 emotion dataset~\cite{goodfellow2013challenges}. We combine these results into a multi-channel model to classify emotion into four categories \textit{(happy, sad, angry, neutral)}.

We combine the time-varying emotion estimates of each pedestrian with path prediction for collision-free, socially normative robot navigation. We present a new data-driven mapping, \textbf{TEM} (Trajectory-based Emotion Model), which maps learned emotions to proxemic constraints relating to the comfort and reachability distances~\cite{ruggiero2017effect}. These distances restrict the robot motion and navigation to avoid intruding through pedestrian's peripersonal and interpersonal social spaces. 
The combination of emotional and proxemic constraints improves both social comfort and navigation. We have evaluated the performance of our algorithm: \\
$\bullet$ quantitatively on a dataset of real-world videos consisting of tens of pedestrians, including dense scenarios, where we measured the number of proxemic intrusions our robot avoided, and \\
$\bullet$ qualitatively in a lab setting with a Pepper humanoid robot and a total of $11$ pedestrians with real-world intentions. Our subjects felt comfortable in the environment, and they could perceive the robot's subtle reaction to their emotion.

The rest of the paper is organized as follows. Section II provides an overview of related work. We introduce the terminology and present our algorithm to model emotional constraints and use them for path prediction in Section III. In Section IV, we present our socially-aware robot navigation scheme. We highlight our algorithm's performance on benchmarks and describe results in Section V.

\section{Related Work}
\label{sec:RelatedWork}
In this section, we discuss previous robot navigation algorithms that focus on physical and social constraints. We also review related work on emotion detection from facial expressions and trajectories.

\subsection{Physical Navigation}
Prior work on robot navigation in pedestrian environments mainly focused on solving physical constraints such as collision avoidance. Many systems for robot navigation in urban environments implemented robots that autonomously navigated previously-mapped urban paths in the presence of crowds~\cite{morales2009autonomous,kummerle2015autonomous}. Bauer et al.~\cite{bauer2009autonomous} created a robot for navigating urban environments without GPS data or prior knowledge. Fan et al.~\cite{fan2019getting} proposed a navigation framework that handles the robot freezing, and the navigation lost problems. The collision-avoidance approaches used by these navigation algorithms include potential-based approaches for robot path planning in dynamic environments~\cite{Pradhan11_potential_function}, probabilistic methods or Bayesian velocity-obstacles methods~\cite{Fulgenzi07_PVO,kim2015brvo} for pedestrian trajectory prediction. Other methods include a partially-closed loop receding horizon control~\cite{DuToit10} for non-conservatively avoiding collisions with dynamic obstacles. Faisal et al.~\cite{faisal2018human} tuned fuzzy logic controllers with four different methods to identify which algorithm minimizes the travel time to a goal destination. Recently, learning-based collision avoidance approaches have been introduced for collision avoidance navigation including both supervised and unsupervised learning~\cite{sergeant2015multimodal,ross2013learning,pfeiffer2017perception,kahn2017uncertainty,fan2018crowdmove}.

\subsection{Socially-aware Robot Navigation}

Humans navigating among crowds follow social norms relating to the speed of movement or personal space. In particular, we tend to observe the emotions displayed by others and adjust our paths in response. Correspondingly, there is much prior work on having mobile robots navigate among humans in a socially-aware manner~\cite{pandey2010framework,kruse2013human,okal2016learning,ferrer2013robot,kuderer2012feature}. Some navigation algorithms generate socially compliant trajectories by predicting the pedestrian movement and forthcoming interactions~\cite{kuderer2012feature} or use modified interacting Gaussian processes to develop a probabilistic model of robot-human cooperative collision avoidance~\cite{Trautman13}. Vemula et al.~\cite{vemula2018social} proposed a trajectory prediction model that captures the relative importance of each pedestrian in a crowd during navigation.
Other methods tend to model interactions and personal space for human-aware navigation~\cite{barnaud2014proxemics} or use learning-based approaches to account for social conventions in robot navigation~\cite{luber2012socially,Higueras2014Robot,Vigo2014human}. Many explicit models have been proposed for social constraints to enable person-acceptable navigation~\cite{sisbot2007human,kirby2009companion}. However, these methods do not take into account psychological constraints or emotions of the pedestrians.

\subsection{Emotion Characteristics from Faces}
In recent years, research in computer vision and AI has focused on emotion identification from facial expressions. Most of these methods use neural-network based approaches to identify emotions trained on popular datasets such as FER~\cite{goodfellow2013challenges}. Liu et al.~\cite{liu2014facial} presented a novel Boosted Deep Belief Network (BDBN) based three-stage training model for facial expression recognition. An annotation method, EmotioNet, predicted action units and their intensities as well as emotion category for a million facial expressions in the wild~\cite{fabian2016emotionet}. A dataset, EMOTIC, was proposed to facilitate emotion recognition given the context of the environment~\cite{kosti2017emotion}. A recent survey on state-of-the-art methods for recognizing emotion characteristics from facial expressions is given in~\cite{rouast2019deep}. In this paper, we use a convolutional neural network based on the Xception~\cite{chollet2017xception} for emotion recognition based on facial expressions.

\subsection{Emotion Characteristics from Trajectory}
The direction and speed with which people move help to predict future behavior, including both pedestrians' behavior and their emotional reactions~\cite{Sartori2011}.  For example, people likely walk slower when they feel depressed, walk in a less direct path if they are distracted, and may change both speed and direction if they are uncertain or ambivalent about their path~\cite{Ostir2000}. A clear benefit in using trajectory tracking as a way of predicting future behavior is that it is a relatively implicit measure. People are not generally aware of the information that their trajectory may be conveying, so this channel tends to have relatively high fidelity across settings and can be used for behavior or emotional classification. There is minimal work on modeling emotions from trajectories. Our work is the first approach that combines information from the trajectory channel with facial expressions to predict emotions and use them for socially-aware robot navigation.

\section{Emotion Learning}
We propose a joint pedestrian emotion-model from trajectories and faces. In this section, we first define an emotion state, then introduce our notation, and give an overview of our approach.

\subsection{Emotion State}
\begin{figure}[t]
    \centering
    \includegraphics[width=\linewidth]{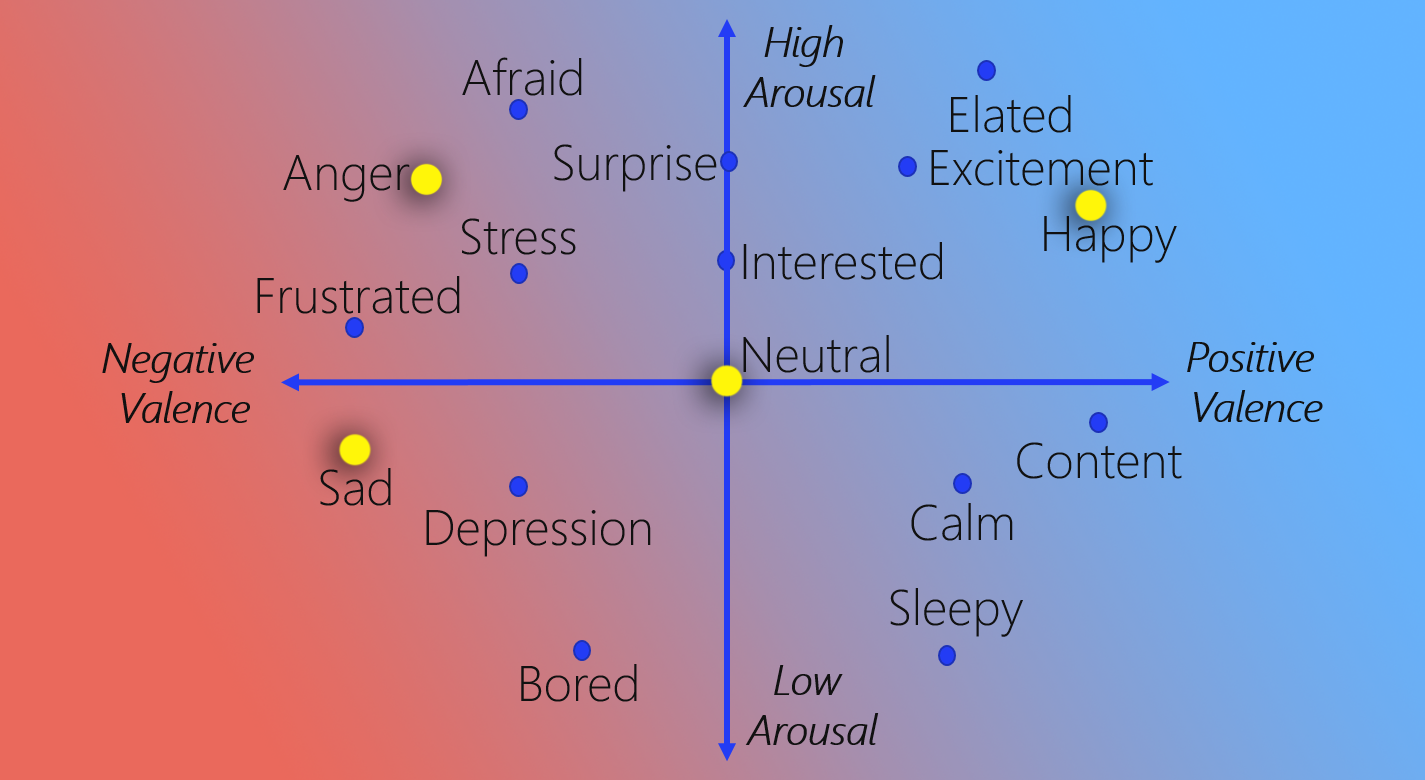}
    \caption{\textbf{Affect Space and Discrete Emotions}: All discrete emotions can be represented by points on a 2D affect space of Valence and Arousal from the PAD model~\cite{loutfi2003social,ekman1967head}.}
    \label{fig:affectSpace}
    \vspace{-15pt}
\end{figure}

Most of the previous literature has modeled emotions as either discrete categories or as points in a continuous space of emotion dimensions. Discrete categories include basic emotions such as anger, disgust, fear, joy, sadness, and surprise as well as other emotions such as pride, depression, etc. On the other hand, Ekman and Wallace~\cite{ekman1967head} used ``affects" to represent emotions. \textit{Affect} is a key characteristic of emotion and is defined as a 2-dimensional space of (1) valence, the pleasure-displeasure dimension; and (2) arousal, the excited-sleep dimension. All discrete emotions can be represented by points in this 2D affect space (Fig~\ref{fig:affectSpace}). In this paper, we use the pleasure and arousal dimensions from the PAD model and use four basic emotions (\textit{happy, angry, sad, neutral}) representing emotional states that last for an extended period and are more abundant during walking~\cite{ma2006motion}. These four emotions capture the spectrum of the emotion space, and a combination of them can be used to represent other emotions~\cite{mikels2005emotional}.

\subsection{Notation}
We introduce the terminology and symbols used in the rest of the paper. We refer to an agent in the crowd as the {\em pedestrian} whose {\em state} includes his/her emotion characteristics. This state, denoted by the symbol $\vec{x}_p$, governs the pedestrian's position on the ground plane and facial features:
\begin{equation}\label{eqn:stateVector}
\vec{x_p}=[\vec{p}_p \; \vec{v}_p^c \; \vec{f}\; \vec{v}_p^{pred}\;  \vec{E}^{f}\;  \vec{E}^{t} ]^\mathbf{T};
\end{equation}
where $\vec{p} \in \mathbb{R}^2$ is the pedestrian's position which is used to compute emotion from the trajectory; $\vec{v}^c \in \mathbb{R}^2$  is his/her current velocity; and $\vec{v}^{pred} \in \mathbb{R}^2$ is the {\em predicted velocity} on a 2D ground plane. A pedestrian's current velocity $\vec{v}^c$ tends to be different from their optimal velocity (defined as the predicted velocity $\vec{v}^{pred}$) that they would take in the absence of other pedestrians or obstacles in the scene to achieve their intermediate goal. $\vec{f}$  is their face pixels re-aligned to $48 \times 48$ which is used to compute the facial emotion. $\vec{E}^f \in \mathbb{R}^3$ and $\vec{E}^t \in \mathbb{R}^3$ are their facial and trajectory emotion vectors. The union of the states of all the other pedestrians and the current positions of the obstacles in the scene is the current state of the environment denoted by the symbol $\mathbf{X}$. 

We do not explicitly model or capture pairwise interactions between pedestrians. However, the difference between $\vec{v}^{pred}$ and $\vec{v}^c$ provides partial information about the local interactions between a pedestrian and the rest of the environment.
Similarly, we define the robot's state, $\vec{x_r}$, as
\begin{equation}\label{eqn:stateVectorRobot}
\vec{x_r}=[\vec{p}_r \; \vec{v}_r^c \; \vec{v}_r^{pref} ]^\mathbf{T};
\end{equation}
where $\vec{v}^{pref}$ is the {\em preferred velocity} of the robot, defined as the velocity the robot would take based on the present and predicted position of the pedestrians and the obstacles in the scene.

We represent the emotional state of a pedestrian by a vector $\vec{E}^t = [h, a, s]$, where $h, a$, and $s$ correspond to a scalar value of happy, angry, and sad emotions (normalized to $[0, 1]$), respectively. Using the $\vec{E}^t$, we can also obtain a single emotion label $e$ as follows:
\begin{eqnarray}
    e = 
\begin{cases}
    happy, & \text{if } (h > a) \land (h > s) \land (h  > \theta)\\
    angry, & \text{if } (a > h) \land (a > s) \land (a  > \theta)\\
    sad, & \text{if } (s > h) \land (s > a) \land (s  > \theta)\\
    neutral,              & \text{otherwise}
\end{cases} \label{eq:emLabel}
\end{eqnarray}
where $\theta$ is a scalar threshold. In this paper, we use an experimentally determined value of $\theta = 0.55$.

\subsection{Overview}
We present an overview of our approach in Figure~\ref{fig:overview}. Our method takes a streaming video as input from two camera channels, a fixed, overhead camera, and an onboard robot camera. We perform a large-scale Mechanical Turk study on a crowd dataset to establish a linear mapping \textit{(TEM)} between the motion model parameters and pedestrian emotion using multiple linear regressions after we obtained the labels using a perception user study. Later we use this mapping and compute trajectory-based emotions for the pedestrians. We also use a fully-convolutional neural network (which has been trained on the FER-2013 emotion dataset~\cite{goodfellow2013challenges}) to compute the facial expression based-emotions. We combine these multi-channel emotions along with proxemic constraints and a collision-avoidance algorithm to perform socially-aware robot navigation.


\section{Emotion Learning}
In this Section, we describe our joint pedestrian emotion model that combines emotion learning from trajectories and facial features.

\subsection{Emotion Learning from Trajectories (TEM)}\label{sec:trajEmotion}
\vspace{-5pt}
Our goal is to model the levels of perceived emotion from pedestrian trajectories. 
We use a data-driven approach to model pedestrians' emotions. We present the details of our perception study in this section and derive the trajectory-based pedestrian emotion model (TEM) from the study results.

\subsubsection{Study Goals}
This web-based study is aimed at obtaining the emotion labels for a dataset of pedestrian videos. We use a 2D motion model~\cite{van2011reciprocal} based on reciprocal velocity obstacles (RVO) to model the motion of pedestrians. We obtained scalar values of perceived emotions for different sets of motion model parameters.

\begin{figure}[t]
  \includegraphics[width =0.99\linewidth]{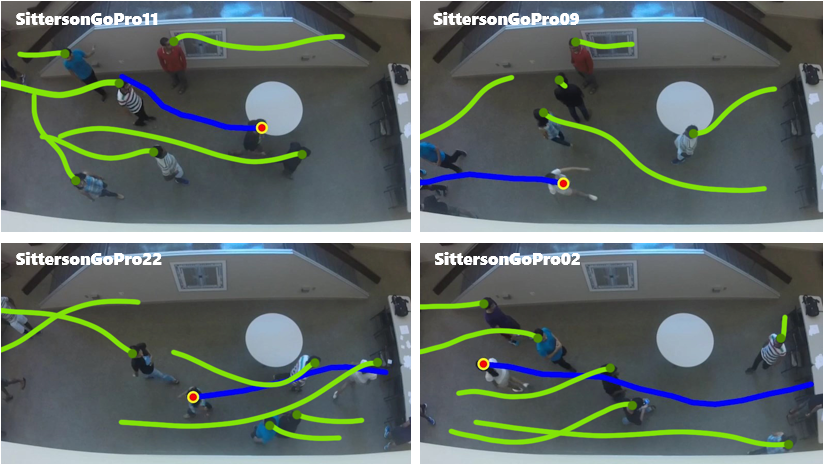}
  \caption{To compute a data-driven emotion mapping, we collected $23$ videos of pedestrians walking in a corridor for our Mechanical Turk perceptual user study. Users were asked to label the emotion of one pedestrian (marked in \textcolor{blue}{blue}).}
  \vspace{-15pt}
  \label{fig:trajVideo}
\end{figure}

\subsubsection{Participants}
We recruited $100$ participants ($77$ male, $23$ female, $\bar{x}_{age}$ = $33.24$, $s_{age}$ = $7.81$) from Amazon MTurk to answer questions about a dataset of simulated videos. 

\subsubsection{Dataset}
We collected $23$ videos of pedestrians walking in a corridor. In each video, a single pedestrian was highlighted by a circle and his/her trajectory (Figure~\ref{fig:trajVideo}). We computed the motion model parameters of the pedestrians using a Bayesian learning approach~\cite{kim2015brvo}. The motion model corresponds to the local navigation rule or scheme that each pedestrian uses to avoid collisions with other pedestrians or obstacles. In particular, we consider the following motion parameters for each pedestrian:
\begin{itemize}
\item Planning Horizon (how far ahead of the agent plans),
\item Effective Radius (how far away an agent stays from other agents), and
\item Preferred Speed.
\end{itemize}

We represent these motion model parameters as a vector ($\vec{P}\in\mathbb{R}^3$): \textit{Planning~Horiz, Radius, Pref~Speed}. Table~\ref{tab:dataRange} shows the range of the values of the parameters used. These values cover the range of values observed in the real world. We will release this dataset and the motion model parameters.


\begin{table}[ht]
\centering
\begin{tabular}{|l|c|c|c|c|}
\hline
\multicolumn{1}{|c|}{Parameter (unit)} & Min & Max & Average & Variance \\ \hline
Planning~Horiz (s)                     & 0.09   & 2.21  & 1.25   & 0.57  \\ \hline
Radius (m)                             & 0.30 & 0.92   & 0.61  & 0.05  \\ \hline
Pref~Speed (m/s)                       & 0.93 & 2.33 & 1.39  & 0.11 \\ \hline
\end{tabular}
\caption{\textbf{Values of Motion Parameters}: We present the range and average values of motion parameters obtained from the dataset. }
\vspace{-12pt}
\label{tab:dataRange}
\end{table}

\subsubsection{Procedure}
In the web-based study, participants were asked to watch a random subset of $8$ videos from the dataset. Participants then answered whether the highlighted agent was experiencing one of the basic emotions (happy, angry, or sad) on a 5-point Likert scale from \textit{strongly disagree (1) - strongly agree (5)}. Participants were presented the videos in a randomized order and could watch the videos multiple times if they wished. Before completing the study, participants also provided demographic information about their gender and age.


\subsubsection{Analysis}
We average the participant responses to each video to obtain a mean value corresponding to each basic emotion: $V_{h}, V_{s}, V_{a}$ (normalized to $[0, 1]$). Using these values, we obtain the emotion vector $\vec{E} = [V_{h}, V_{s}, V_{a}]$ and the emotion label $e$ using Equation~\ref{eq:emLabel}.

We obtain emotion vectors $\vec{E}^t_i$ corresponding to each variation of the motion model parameters $\vec{P}_{i}$ for the $23$ data points corresponding to $23$ videos in the simulated dataset. We use this labeled data to fit a model for emotion computation using multiple linear regression. We chose linear regression because it is computationally inexpensive and easily invertible. Other forms of regressions can also be employed. TEM takes the following form:
\begin{small}
\begin{equation} \label{eq:mapping}
\vec{E}^t  = \begin{pmatrix}
-0.15 &    0.00 &    -0.12 \\
0.24 &    -0.61 &    0.20 \\
-0.02 &    0.79 &    0.11\\
\end{pmatrix}*\vec{P}.
\end{equation}
 \end{small}
    \vspace*{-17pt}


We can make several inferences from the values of the coefficients of the mapping between perceived emotion and the motion model parameters. The radius of the pedestrian representation affects the perception of anger negatively and the perception of sadness positively, whereas it doesn't affect the perception of happiness significantly. Therefore, increasing the radius makes pedestrians appear sad and decreasing it makes them appear angry. Similarly, we can control the value of the planning horizon to control the perception of happiness and anger. Increasing the planning horizon makes pedestrians appear angrier whereas decreasing it makes them appear happier. The preferred speed affects the perception of happiness positively and the perception of anger and sadness negatively.

We use the linear model to predict the value of a pedestrian's emotion label given the motion model parameters. We compute the accuracy of TEM using 10-fold cross-validation on our labeled dataset. We perform 100 iterations of the cross-validation and obtain an average accuracy of $85.33\%$ using the actual and predicted emotion values.

\vspace{-5pt}
\subsection{Emotion Learning from Facial Features}
\vspace{-5pt}

In this section, we discuss the architecture of the neural network used to detect faces from a video stream captured by the robot and to classify the emotions for those faces. We leverage a CNN based on the Xception~\cite{chollet2017xception} architecture to predict the emotions, $\vec{E}^f$,  from faces. Our network is fully-convolutional and contains $4$ residual depth-wise separable convolutions where a batch normalization operation and a ReLU activation function follows each convolution. The final layer of our network is followed by a global average pooling and a soft-max activation function. The network has been trained on the FER-2013 dataset, which contains $35,887$ grayscale images. We choose images that belong to one of the following classes: {\textit{angry, happy, sad, neutral}}.

We use a fully-convolutional network because each face is localized to a roughly square part of the video frame and detecting its emotion does not require knowledge of the ``global" information about other faces in the frame. We use depth-wise separable convolutions since they need fewer numerical computations on images with a depth of more than 1 - here, we are dealing with RGB video, so it has three input channels. For more details, we refer the readers to~\cite{chollet2017xception}.

\subsection{Joint Pedestrian Emotion Model}
\label{sec:jointEmotion}
\vspace{-5pt}
We combine the computed emotions ($\vec{E}^{t}$ and $\vec{E}^{f}$) using a reliability weighted average. Since facial features are more unreliable (faces are partially visible or far away from the camera), we define the joint pedestrian emotion as: 
\begin{eqnarray}
\vec{E} = \frac{\alpha \vec{E^t} + \left \lfloor{\max (E^f) + 1/2}\right \rfloor \vec{E^f}}{\alpha + \left \lfloor{\max (E^f) + 1/2}\right \rfloor} 
\label{eq:jointEmotion}
\end{eqnarray}
where $\alpha \in [0,1]$ is the pedestrian tracking confidence metric based on~\cite{bera2014adapt}. Based on the unreliability in the facial features, Equation~\ref{eq:jointEmotion} computes a weighted average of the emotions predicted from faces and trajectories. Whenever facial emotion is unavailable, we use $\vec{E} = \vec{E^t}$. We also compute the emotion label $e$ using Equation~\ref{eq:emLabel} from $\vec{E}$.
\section{Proxemic Constraints and Path Prediction}
\vspace{-5pt}

We present our real-time approach that computes a combination of physical and social constraints and uses them for navigation (Figure~\ref{fig:overview}). Most of the prior work in robot navigation is limited to collision avoidance while accounting for kinematics and dynamic constraints. However, when navigating in an environment alongside humans, additional social and emotional constraints such as proximity must also be respected. Depending on the emotions or behaviors of the humans in the environment, such constraints should also be satisfied in addition to the physical constraints. 

Given a pedestrian tracker~\cite{bera2014real,bera2014rcrowdt} that works well on low- to medium- density crowds, we extract pedestrian trajectories. We use Bayesian inference to compensate for noise in the trajectories extracted by the pedestrian trackers. Both the sensor error and the prediction error are assumed to follow a zero-mean Gaussian distribution. We use an Ensemble Kalman Filter (EnKF) and Expectation Maximization (EM)~\cite{kim2015brvo} to estimate the most-likely motion model parameters $\vec{P}$ of each pedestrian. Using these parameters, we learn the pedestrians' emotions using TEM (Equation~\ref{eq:mapping}). 

We formulate emotion-based proximity constraints and combine them with collision-avoidance constraints for robot navigation. Pedestrians' emotion predictions are also used for path prediction and a socially-aware robot navigation algorithm.

\vspace{-5pt}
\subsection{Pedestrian Path Prediction}
\vspace{-5pt}
\label{sec:pathprediction}
Our pedestrian path prediction algorithm takes the state of all pedestrians in the environment $\mathbf{X}$ for the previous $n$ timesteps at time $t$ and predicts the path of a pedestrian $i$ for a future time, $\vec{x}_i^{t + \Delta t}$. This path is predicted in the form of their motion parameters $\vec{P}_i$ because it is the best estimator of their motion in a dense environment. 

The estimated motion parameters may vary slightly during the duration of motion of each pedestrian. We model these variations by an upper bound $\vec{P}_{ub}$ and a lower bound $\vec{P}_{lb}$. We compute the values of $\vec{P}_{ub}$ and $\vec{P}_{lb}$ using the values of the emotion vector $\vec{E}$ using TEM (Equation~\ref{eq:mapping}). Using the computed emotion label $e$ from Equation~\ref{eq:emLabel}, we compute $\vec{P}_{ub}$ by adding $\gamma$\% to the emotion value corresponding to $e$ and adding $(\gamma/3)$\% to the other traits. Similarly, we compute $\vec{P}_{lb}$ by subtracting $\gamma$\% from the emotion value corresponding to $e$ and subtracting $(\gamma/3)$\% from the other traits. The users can control the value of $\gamma$. In this paper, we use a value of $\gamma = 5\%$ that captures the noise and natural variance of the pedestrians' emotions.


 
We assume that for the duration of the navigation of the robot around a pedestrian, the emotion of that pedestrian does not dramatically change and lies within a range of variance. Under these assumptions, a substantial change in the emotion vector at successive timesteps indicates an error in the prediction. To account for this error, we clamp the estimated motion model parameters to the corresponding boundary value if they are out of the bounds $\vec{P}_{lb}$ and $\vec{P}_{ub}$. Otherwise, we use them directly for path prediction. We recompute the emotion vector $\vec{E}$ according to the updated motion model parameters. 
 \vspace*{-8pt}
\begin{small}
\begin{equation}
    \vec{P}= 
\begin{cases}
    \vec{P}_{lb},& \text{if } \vec{P}_{i} \leq \vec{P}_{lb_i},   \forall i \\
    \vec{P}_{ub},& \text{if } \vec{P}_{i} \geq \vec{P}_{ub_i},   \forall i \\
    \vec{P},              & \text{otherwise}
\end{cases}
\end{equation}
 \end{small}
    \vspace*{-17pt}

We use these motion parameters $\vec{P}$ to perform path prediction using the GLMP algorithm~\cite{beraglmp}, which computes various local movement patterns of pedestrians as well as the crowd's overall global movement patterns.

\section{Emotionally Intelligent Robot Navigation}
\vspace{-5pt}
We use the emotions computed using Equation~\ref{eq:jointEmotion} to perform socially-aware robot navigation using proxemic distances as described below.
\vspace{-10pt}
\subsection{Peripersonal and Interpersonal Social Spaces}
\label{sec:proxemicdistances}
\vspace{-7pt}
Recent neuro-cognitive studies~\cite{ruggiero2017effect} have suggested a relationship between  peripersonal-action (reachability distance) and interpersonal-social (comfort distance). In particular, reachability distance refers to the distance at which pedestrians feel comfortable interacting with other pedestrians, and comfort distance refers to the distance at which pedestrians feel comfortable with the presence of a pedestrian. These proxemic behaviors offer a window into everyday social cognition by revealing pedestrians’ emotional states and responses.

To enable the robot to perform socially-aware navigation, we incorporate these proxemic distances in the navigation algorithm. We compute the numerical values of {\em comfort distance ($cd_e$)} and {\em reachability distance ($rd_e$)} to perform socially-aware navigation (where $e$ indicates the computed emotion label, Section~\ref{sec:jointEmotion}). Though these distances depend on cultural norms, environment, or an individual's personality, we mainly focus on variations originating from the differences in emotions. Specifically, we exploit the experiments described in~\cite{ruggiero2017effect} to compute the limits on an individual's comfort and reachability distances (Table~\ref{tab:extradist}). 
\vspace{-8pt}
\begin{table}[h]
\centering
\begin{tabular}{|c|c|c|}
\hline
          & \textbf{\textit{Comfort Distance }($cd_e$)} & \textbf{\textit{Reachability Distance} ($rd_e$)} \\ \hline
\textit{Happy} &  90.04        & 127.38          \\ \hline
\textit{Sad} &  112.71          & 148.97          \\ \hline
\textit{Angry} &  99.75          & 138.38          \\ \hline
\textit{Neutral} &  92.03          & 136.09          \\ \hline
\end{tabular}
\caption{\textbf{Peripersonal and Interpersonal Social Spaces:} The comfort and reachability distances indicate the minimum distance at which the pedestrian feels uncomfortable with the robot~\cite{ruggiero2017effect}. The distances are given in \textit{cm}.}
\vspace{-15pt}
\label{tab:extradist}
\end{table}

 \begin{figure}[t]
  \centering
  \includegraphics[width =\linewidth]{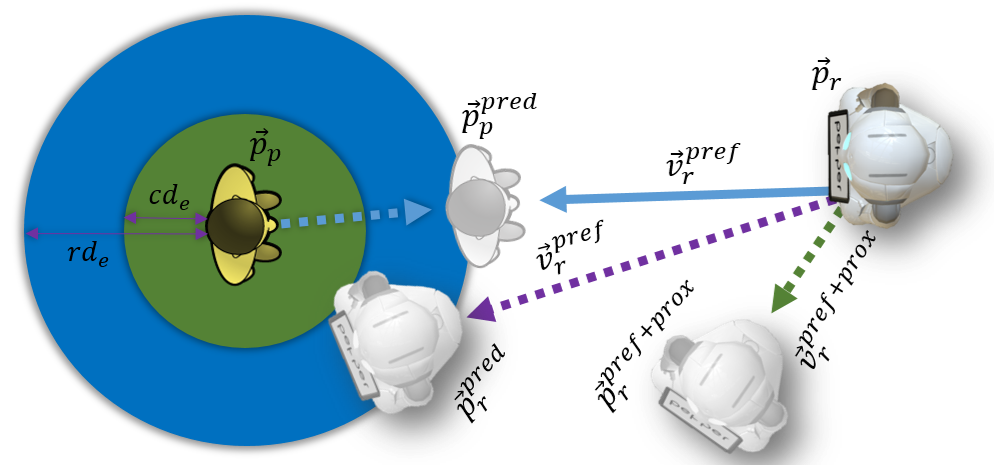}
  \caption{Our robot navigation algorithm satisfies the proxemic distance constraints, including peripersonal space (\textcolor{green}{green}) and interpersonal space (\textcolor{blue}{blue}). The trajectory computed by our algorithm does not intrude onto these spaces, whereas a robot that fails to consider the reachability distance (\textcolor{violet}{purple} trajectory) may cause discomfort to some pedestrians.}
  \label{fig:navigation}
  \vspace{-15pt}
\end{figure}


\subsection{Socially-Aware Robot Navigation}
\vspace{-7pt}
We present an extension to Generalized Velocity Obstacles (GVO)~\cite{WilkieGVO, bera2017sociosense} that takes into account the comfort ($cd_e$) and reachability ($rd_e$) distances and enables socially-aware collision-free robot navigation through a crowd of pedestrians. The GVO algorithm uses a combination of local and global methods, where the global metric is based on a roadmap of the environment, and the local method computes a new velocity for the robot. We compute the comfort and reachability distances using the emotional labels $e$ (Section~\ref{sec:jointEmotion}) and use them in the computation of the new velocity for the robot (i.e., the local method of the GVO algorithm). In this novel formulation, we also take into account the dynamic constraints of the robot. Even though our socially-aware navigation algorithm is illustrated with GVO, our approach is agnostic to the underlying navigation algorithm and can be combined with other methods like potential field methods.

We provide an illustration of a robot (located at $\vec{p}_{r}$) avoiding a collision with an oncoming pedestrian (located at position $\vec{p}_{p}$) that uses the comfort and reachability distances in Figure~\ref{fig:navigation}. The pedestrian's two associated proxemic distances are the comfort distance of $cd_e$ and the reachability distance of $rd_e$. At this time instant, the global navigation method has computed a preferred velocity $\vec{v}_{r}^{pref}$ for the robot that would navigate it to its goal position in the absence of any static or dynamic obstacles. Using the path prediction algorithm described in Section~\ref{sec:pathprediction}, the robot predicts that the pedestrian will move to the position $\vec{p}_{p}^{pred}$ during the next time window. Using this information, the robot computes a new velocity such that it avoids a collision with the pedestrian. However, this method doesn't account for the pedestrian's proxemic distances and is not sufficient for socially-aware navigation. Therefore, the robot alters its goal position and velocity in a way that takes into account both comfort and reachability: $\vec{p}_{r}^{pref+prox}$ and $\vec{v}_{r}^{pref+prox}$. This updated velocity $\vec{v}_{r}^{pref+prox}$ successfully accounts for the pedestrian's comfort and reachability distances, whereas the velocity $\mathbf{v}_{r}^{pref}$ results in the robot intruding on the pedestrian's comfort distance. During navigation, our algorithm avoids any steering inputs that result in a collision with the predicted pedestrian positions.

\vspace{-6pt}
\section{Performance and Analysis}
\label{sec:Experiments}
\vspace{-5pt}

\begin{figure}[h]
  \centering
  \includegraphics[width =0.99\linewidth]{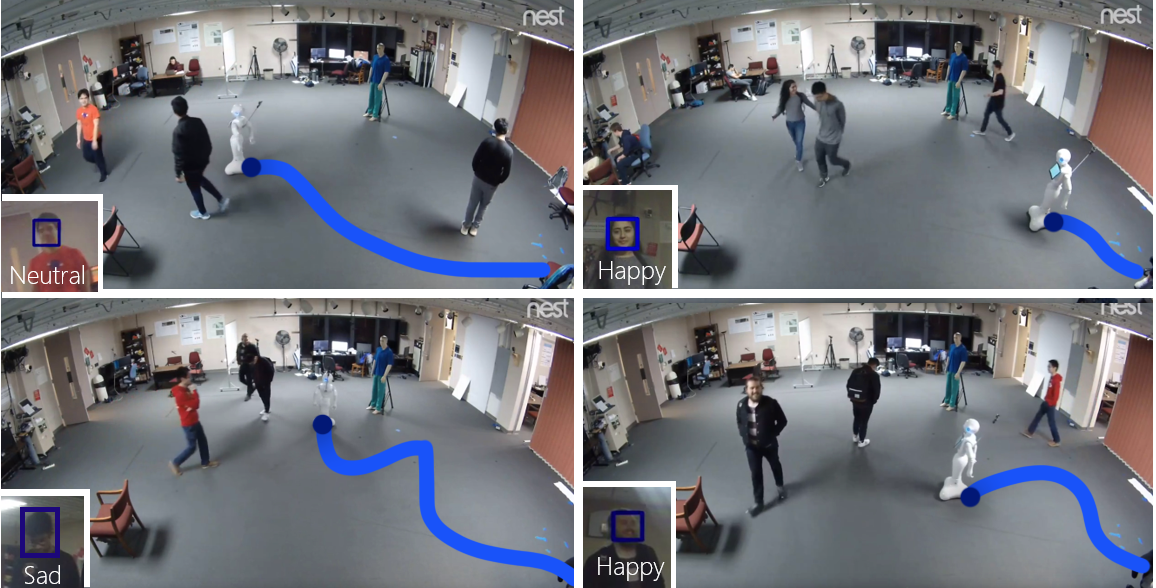}
  \vspace*{-1em}
  \caption{The robot learns pedestrians' emotions and their proxemic constraints in real-time. These distances restrict the robot motion and navigation to avoid intruding on pedestrian's peripersonal and interpersonal social spaces. The combination of emotional and proxemic constraints improves both social comfort and navigation.}\label{fig:results}
  \vspace*{-1em}
\end{figure}

We have implemented our algorithm on a semi-humanoid robot, Pepper. It is about $1.2m$ tall with a top speed of $0.83m/s$ and an on-board camera with $2592x1944$ Active Pixels. We conducted experiments in a lab setting (Fig.~\ref{fig:results}). We recruited $11$ participants and asked them to assume that they are experiencing a certain emotion and walk accordingly. Previous studies show that non-actors and actors are both equally good at walking with different emotions~\cite{roether2009critical}. We do not make any assumptions about how accurately the subjects acted or depicted the emotions.  The participants reported being comfortable with the robot in the scene. Participants with \textit{sad} emotions reported being given a wider space to walk by the robot. Participants with \textit{angry} emotions reported that the robot made way for the pedestrian more swiftly. Participants with \textit{happy} and \textit{neutral} emotions didn't report significant changes, but some noticed a minor slow down in the robot's speed.

We also quantitatively evaluate the performance of our socially-aware navigation algorithm with GVO~\cite{WilkieGVO}, which does not take into account proxemic or emotional constraints. We compute the number of times the non-social robot intrudes on the peripersonal and the interpersonal spaces of the pedestrians, thereby resulting in emotional discomfort. We also measure the additional time a robot with our algorithm takes to reach the goal position without any intrusions on pedestrians' comfort distances (hard constraint) and reachability distances (soft constraint). Our results (Table~\ref{tab:navperf}) demonstrate that our robot can reach its goal with $<25$\% time overhead while ensuring that the proxemic spaces of the pedestrians aren't violated.

\begin{table}[h]
\centering
\begin{tabular}{|c|c|c|c|}
\hline
\textit{\textbf{Dataset}} & \textit{\textbf{Additional Time}} & \textit{\textbf{Performance}} & \textit{\textbf{Intrusions Avoided}} \\ \hline
\textit{NDLS-1}           & 19.44\%                              & 2.89E-04 ms                             & 31                                   \\ \hline
\textit{NDLS-2}           & 21.08\%                              & 2.12E-04 ms                             & 26                                   \\ \hline
\textit{NPLC-1}         & 16.71\%                              & 2.29E-04 ms                             & 30                                   \\ \hline
\textit{NPLC-3}       & 18.93\%                              & 3.09E-04 ms                            & 22                                   \\ \hline
\textit{UCSD-Peds1}       & 24.89\%                              & 3.51E-04 ms                              & 11                                   \\ \hline
\textit{Students}         & 9.12\%                              & 0.78E-04 ms                             & 16                                   \\ \hline
\textit{seq\_hotel}       & 11.89\%                              & 1.07E-04 ms                             & 9                                   \\ \hline
\textit{Street}       & 11.09\%                              & 1.27E-04 ms                              & 13                                   \\ \hline

\end{tabular}
\caption{\textbf{Navigation Performance:} A robot using our socially-aware navigation algorithm can reach its goal position (within $\tilde 1m$ accuracy), while ensuring that the peripersonal/interpersonal space of any pedestrian is not intruded on with $<25$\% overhead. We evaluated this performance in a simulated environment, though the pedestrian trajectories were extracted from the original video. }
\label{tab:navperf}
\vspace{-15pt}
\end{table}

\section{Conclusions, Limitations, and Future Work}
\vspace{-5pt}
We present a real-time data-driven planning algorithm that takes the emotional state of the pedestrians into account to perform socially-aware navigation. We predict pedestrians' emotions based on the PAD model using information from two different channels of expression: faces and trajectories. We extract the trajectory of each pedestrian from a video stream and use Bayesian learning algorithms to compute his/her motion model and emotional characteristics. The computation of this trajectory-based emotion model (\textbf{TEM}) is based on the results of a perception user study that provides emotion labels for a dataset of walking videos. We also compute the facial expression-based emotion using a CNN classifier trained on the FER-2013 emotion dataset~\cite{goodfellow2013challenges}. Our work is the first approach that combines information from the trajectory channel with facial expressions to predict emotions and use them for socially-aware robot navigation.

Our approach has some limitations. We assume that the pedestrian trajectories are captured from an overhead camera and the on-board robot camera captures the facial expressions. Both of these channels have to be perspective corrected. The emotion classification is based on the PAD model, and we choose only four emotions, i.e., \textit{happy, sad, angry and neutral}. These may not be sufficient to understand and capture the observed emotion range. For future work, we would like to learn emotions from full-body gaits and integrate the three sensors channels. We would also want to take group behaviors and cultural norms into consideration for socially-aware navigation. We would also like to compare with an alternate representation of emotion like the Circumplex model~\cite{russell1980affect} to show that this work could be applied to different emotion representations.

\vspace{-10pt}

\bibliographystyle{plain}
\bibliography{ref}
\end{document}